\documentclass[10pt,twocolumn,letterpaper]{article}

\usepackage{iccv}
\usepackage{times}
\usepackage{epsfig}
\usepackage{graphicx}
\usepackage{amsmath}
\usepackage{epstopdf}
\usepackage{amssymb}
\usepackage{bm}
\usepackage{multirow}
\graphicspath{{./images/}}
\usepackage{comment}
\usepackage{subfig}

\usepackage[pagebackref=true,breaklinks=true,letterpaper=true,colorlinks,bookmarks=false]{hyperref}

\iccvfinalcopy % *** Uncomment this line for the final submission

\def\iccvPaperID{1067} % *** Enter the ICCV Paper ID here

\ificcvfinal\fi
\begin{document}

%%%%%%%%% TITLE
%\title{Dual Directed CapsNet via Assisted Center Loss and Targeted Reconstruction}
\title{Dual Directed Capsule Network for Very Low Resolution Image Recognition}

\author{Maneet Singh, Shruti Nagpal, Richa Singh, and Mayank Vatsa\\
IIIT-Delhi, India\\
{\tt\small \{maneets, shrutin, rsingh, mayank\}@iiitd.ac.in}
%\and
}

\maketitle
%\thispagestyle{empty}

%%%%%%%%% ABSTRACT
\begin{abstract}

Very low resolution (VLR) image recognition corresponds to classifying images with resolution $16\times16$ or less. Though it has widespread applicability when objects are captured at a very large stand-off distance (e.g. surveillance scenario) or from wide angle mobile cameras, it has received limited attention. This research presents a novel Dual Directed Capsule Network model, termed as DirectCapsNet, for addressing VLR digit and face recognition. The proposed architecture utilizes a combination of capsule and convolutional layers for learning an effective VLR recognition model. The architecture also incorporates two novel loss functions: (i) the proposed HR-anchor loss and (ii) the proposed targeted reconstruction loss, in order to overcome the challenges of limited information content in VLR images. The proposed losses use high resolution images as auxiliary data during training to ``direct" discriminative feature learning. Multiple experiments for VLR digit classification and VLR face recognition are performed along with comparisons with state-of-the-art algorithms. The proposed DirectCapsNet consistently showcases state-of-the-art results; for example, on the UCCS face database, it shows over 95\% face recognition accuracy when $16\times16$ images are matched with $80\times80$ images. 

\end{abstract}

%%%%%%%%% BODY TEXT
\section{Introduction}

\begin{figure}
     \centering
     \subfloat[][Real-world applications of VLR recognition. Image source: (i) Internet, (ii) UCCS dataset \cite{uccs}]{\includegraphics[width=3.2in]{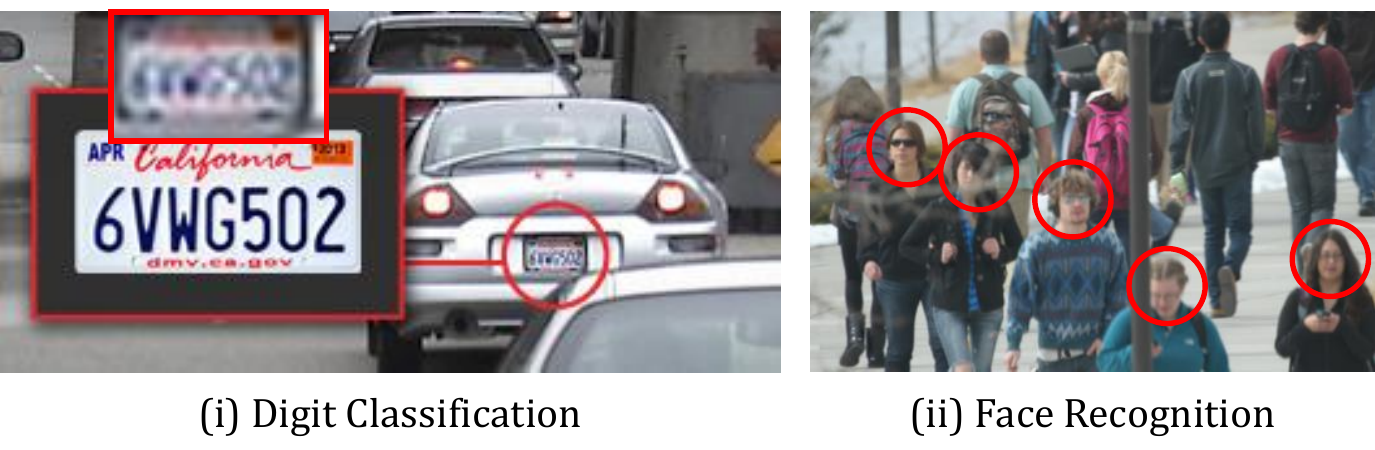}} \\
     \subfloat[][Proposed Dual Directed Capsule Network (DirectCapsNet)]{\includegraphics[width=3.2in]{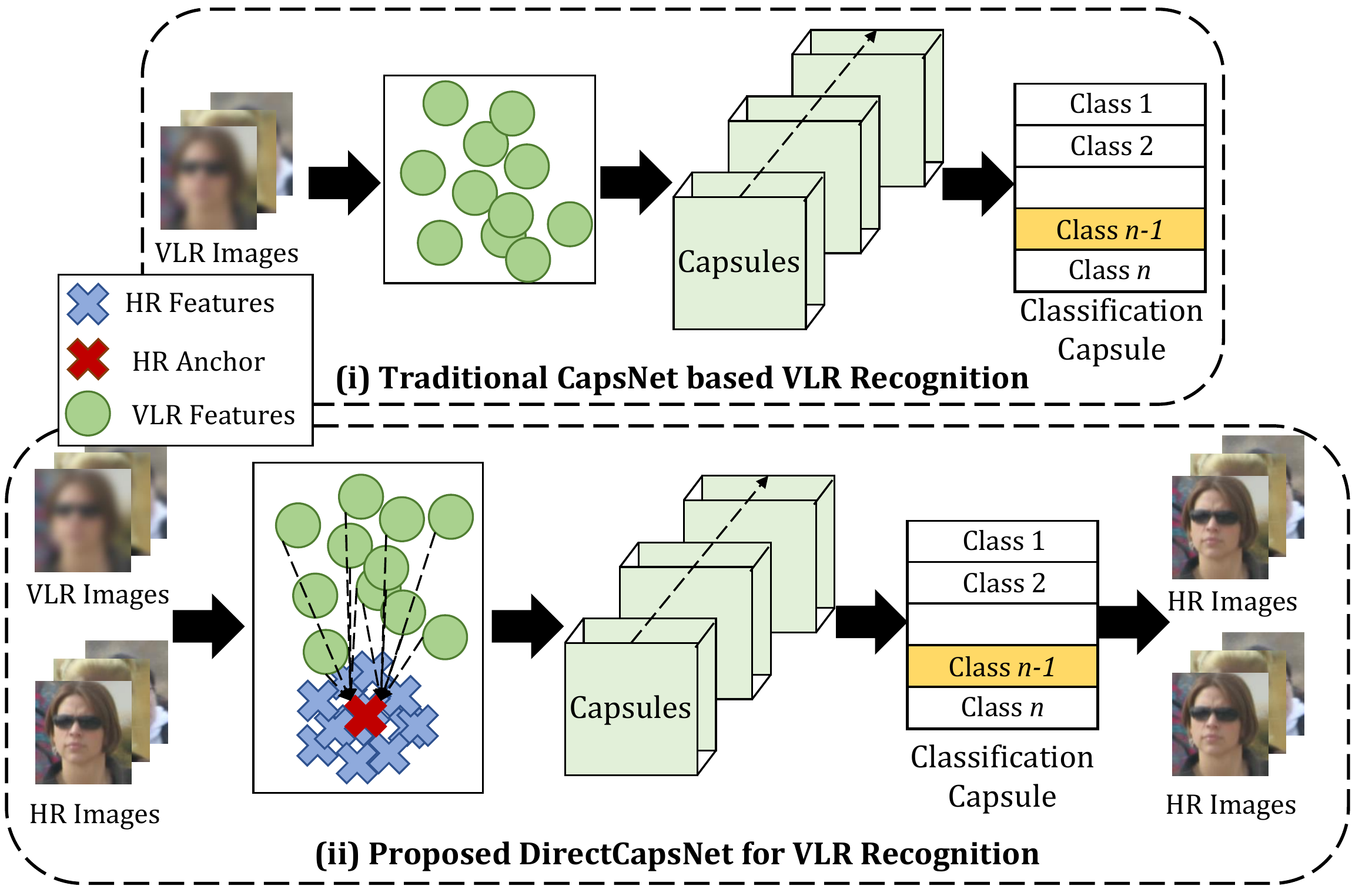}}
     \caption{The proposed DirectCapsNet utilizes HR samples to \textit{direct} learning of more meaningful and discriminative features for VLR image recognition via the proposed HR-anchor loss and the targeted reconstruction loss. }
     \label{fig:intro}
\end{figure}

%The problem of very low resolution (VLR) recognition corresponds to identifying images smaller than $16\times16$ pixels \cite{zou2012very}. It has wide-spread applicability in scenarios of surveillance to monitor humans and objects. 
In typical surveillance scenarios, images are often captured from a large stand-off distance, thus rendering the region of interest to be of a very low resolution (VLR), often times less than $16\times16$ \cite{zou2012very}. Figure \ref{fig:intro}(a) shows sample real-world applications of VLR recognition where the region of interest can be a face, a suspicious object, or the license plate number of a moving vehicle. These samples demonstrate the arduous nature of the problem where some of the key challenges of VLR recognition are the presence of limited information content and blur. VLR recognition also has applicability in image tagging, where multiple objects/people are captured in the frame, and each of these entities are of small resolution.

\begin{comment}
\begin{figure}
\centering
\includegraphics[width=3.2in]{intro.png}
\caption{The proposed Dual Directed CapsNet (DirectCapsNet) utilizes HR samples to \textit{direct} learning of more meaningful and discriminative features for VLR recognition. }
\label{fig:intro}
\end{figure}
\end{comment}

Netzer \textit{et al.} \cite{svhn} demonstrated the poor performance of humans on identifying VLR digits captured in real surroundings. For the Street View House Numbers (SVHN) dataset, the authors observed cent percent accuracy by humans for samples with $101-125$ pixel height. On the other hand, the performance dropped to $82.0\%\pm2\%$ when classifying very low resolution samples, i.e. images of height up to 25 pixels, thereby reinstating the challenging nature of the problem. Direct up-sampling via interpolation could be viewed as a possible solution for VLR recognition, however, multiple studies have demonstrated poor performance owing to the required large magnification factor \cite{li19Tifs,singh18CvprW} and possible introduction of noise, which can also be observed in Figure \ref{fig:intro}(a)(i). Further, in the literature, researchers have also demonstrated the inability of models trained on high resolution (HR) images (containing high information content) to perform well on (V)LR images \cite{singh18CvprW}. The current state of scarce available solutions and the wide applicability of VLR recognition makes it an important problem, demanding dedicated attention.

This research proposes a novel capsule network based model for VLR image recognition. 
%\textit{Dual Directed Capsule Network} (termed as \textit{DirectCapsNet}) for VLR recognition (Figure \ref{fig:intro}(b)). 
Hinton \textit{et al.} \cite{hinton2011transforming} proposed learning ``capsules'', which represent a vector of instantiation parameters in order to encode the input more efficiently. Instantiation parameters may constitute the properties of an image such as the \textit{pose, lighting, and deformation of the visual entity relative to an implicitly defined canonical version of that entity} \cite{hinton2011transforming}. We believe that such parameters would be invariant to the resolution of the image, therefore presenting the potential of being useful for VLR recognition. Due to the limited information content in VLR images, the VLR recognition model could benefit from the information-rich HR samples as well. To this effect, we propose \textit{Dual Directed Capsule Network} (termed as \textit{DirectCapsNet}) (Figure \ref{fig:intro}(b)) to learn meaningful features for VLR recognition, directed (or guided) by the HR samples. 
%via two ways: (i) proposed HR-anchor loss and (ii) proposed targeted reconstruction loss, along with a margin loss for discriminative classification.  
%the proposed network is directed to learn meaningful features by two ways: (i) proposed HR-anchor loss and (ii) proposed targeted reconstruction loss, along with a margin loss for discriminative classification. DirectCapsNet utilizes a capsule network as its base architecture, such that the proposed HR-anchor loss pushes the learned representations of VLR samples towards their HR counter-parts, and targeted reconstructions enforce similar capsule outputs for VLR and HR samples belonging to the same class. 
The contributions of this research are as follows:

%This research proposes \textit{Dual Directed Capsule Network (termed as DirectCapsNet)} for VLR recognition (Figure \ref{fig:intro}(b)). Hinton \textit{et al.} \cite{hinton2011transforming} proposed learning ``capsules'', which represent a vector of instantiation parameters in order to represent the input more efficiently. Since the task of VLR recognition suffers from lack of information content in the input samples, we propose using capsules to efficiently represent VLR input, and utilize the HR samples to guide the VLR samples towards discriminative and information rich features. The proposed DirectCapsNet utilizes HR samples during training to \textit{direct} the features extracted from the VLR samples. To this effect, the proposed network is directed to learn meaningful features by two ways: (i) proposed HR-anchor loss, and (ii) proposed targeted reconstruction loss, along with a margin loss for discriminative classification. DirectCapsNet utilizes a capsule network as its base architecture, such that the proposed HR-anchor loss pushes the learned representations of VLR samples towards their HR counter-parts, and targeted reconstructions enforce similar outputs for VLR and HR samples belonging to the same class. thus, the contributions of this research are as follows:
\begin{itemize}
    \vspace{-4pt}
    \item A novel Dual Directed Capsule Network (\textit{DirectCapsNet}) model is proposed for VLR recognition, which \textit{directs} the features learned from the VLR images containing limited information towards the more meaningful and discriminative features of the HR images.  
    \vspace{-4pt}
    \item Two losses are proposed for directing the VLR recognition model: (i) HR-anchor loss and (ii) targeted reconstruction loss. HR-anchor loss is proposed for the feature learning module, which pushes the VLR features of a particular class towards a representative HR feature (anchor) of that class. Targeted reconstruction loss is utilized at the classification module, where HR images are reconstructed from the capsule outputs of the VLR images, thereby forcing the capsules of VLR and HR images of the same class to be similar. 
    \vspace{-4pt}
    \item Experimental results and analysis demonstrate the advantages of the proposed DirectCapsNet model for VLR digit classification and VLR face recognition. Experiments are performed on the SVHN \cite{svhn}, CMU Multi-PIE \cite{mpie}, and UCCS \cite{uccs} databases, and comparisons are performed with state-of-the-art algorithms. The proposed model yields over 95\% accuracy on the challenging UCCS face database. On the SVHN database, it achieves about 84\% classification accuracy with $8\times8$ VLR images demonstrating an improvement of almost 27\% from the existing results.    
\end{itemize}

\section{Related Work}
There have been several advances in the field of low resolution recognition \cite{jian15,li19Tifs,hetSurvey,Wang2014Review}; however, the area of very low resolution (VLR) recognition remains relatively less explored. As mentioned previously, very low resolution (VLR) recognition refers to identifying regions of interest with $16\times16$ resolution or less. Owing to the limited information content in a given VLR image, a potential solution is to super-resolve or synthesize its higher resolution image \cite{park03super,Wang2014}, which is then used for recognition. While there exists vast literature on super-resolution or synthesis algorithms \cite{Ledig_2017_CVPR, Sajjadi_2017_ICCV, Wang_2018_CVPR}, most of them focus primarily on the visual quality of the generated image, and not on the task of recognition. Zou and Yuen \cite{zou2012very} proposed one of the initial super resolution techniques with specific focus on VLR face recognition. The proposed algorithm utilizes a combination of visual quality based constraint for good quality HR synthesis, and a discriminative constraint for learning features useful for recognition. Singh \textit{et al.} \cite{singh18CvprW} proposed an identity-aware face synthesis technique for generating a HR image from a given LR input. The synthesized images were provided to a Commercial-Off-The-Shelf (COTS) system for recognition.

Apart from super-resolution based techniques, in the literature, researchers have also proposed algorithms for \textit{enhancing} or \textit{improving} the features learned for VLR images by using the information extracted from the HR images. For instance, Bhatt \textit{et al.} \cite{bhatt14Tip} proposed an ensemble-based co-transfer learning algorithm for face recognition. The co-transfer algorithm operates at the intersection of co-training and transfer learning by utilizing the information of HR images for enhancing the VLR classification. Wang \textit{et al.} \cite{wang2016studying} proposed Robust Partially Coupled Networks for VLR recognition. HR images are used as ``auxiliary'' data during training for learning more discriminative information, which might not be available in VLR images. As demonstrated via multiple experiments, using HR images at the time of training, enhances the learned VLR features, resulting in improved recognition performance. Mudunuri and Biswas \cite{mudunuriPami16} proposed a reference-based approach along with multidimensional scaling for learning a common space for HR and VLR images. Recently, Li \textit{et al.} \cite{li19Tifs} analyzed different metric learning techniques for LR and VLR face recognition, by learning a common feature space for HR and LR samples. Ge \textit{et al.} \cite{ge19Tip} proposed a selective knowledge distillation technique for (V)LR face recognition. A base network trained on HR face images is used for selecting the most informative facial features for a (V)LR CNN model, in order to enhance the (V)LR features and the classification performance.

In the literature, VLR recognition algorithms have shown to benefit from HR samples by learning shared representations between the HR and VLR samples \cite{wang2016studying} or by transferring the model information learned by the HR data onto the VLR recognition model \cite{ge19Tip}. By utilizing the additional information from the HR images at the time of training, such algorithms are able to learn more discriminative and meaningful features, as compared to those learned independently from the VLR images. This research proposes to utilize the auxiliary HR samples during training to \textit{direct} the VLR features towards the more informative HR features, via a novel DirectCapsNet model.

%Mudunuri \textit{et al.} \cite{mudunuri18Cvprw} proposed a \textit{two channel Convolutional Neural Network} with Contrastive loss for handling the resolution variations with application to VLR face verification. 

\section{Proposed Dual Directed Capsule Network}
\begin{figure}
\centering
\includegraphics[width=2.8in]{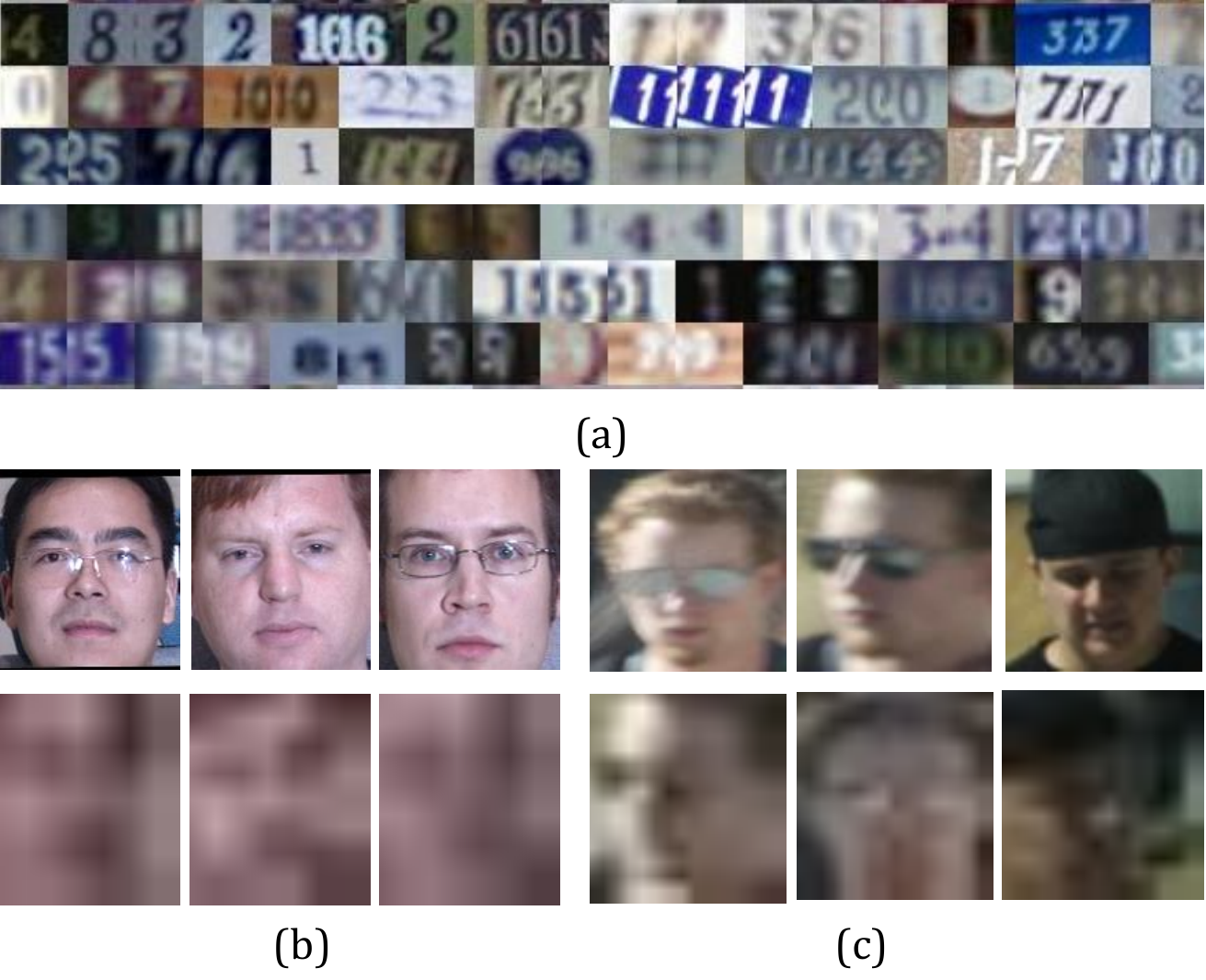}
\caption{Sample HR and VLR images from the (a) SVHN dataset, (b) CMU Multi-PIE dataset, and (c) UCCS dataset. The HR images (first row) contain high information content, which is often missing in the VLR samples (second row). }
\label{fig:vlr_hr}
\vspace{-10pt}
\end{figure}

As shown in Figure \ref{fig:vlr_hr}, the problem of very low resolution (VLR) recognition suffers from the challenge of limited information content in the input images, which often results in lack of discriminative features useful for recognition/classification. In order to overcome this challenge, we propose a novel Dual Directed CapsNet, termed as \textit{DirectCapsNet}. DirectCapsNet enhances the VLR representations by directing them in two ways: via the proposed (i) HR-anchor loss and (ii) targeted reconstruction loss, both of which provide additional supervision using the HR images. The HR information is used to direct/guide the framework to extract discriminative representations even from the VLR images having limited information content. This is accomplished by using the HR-anchor loss which brings the representations of VLR images closer to the representations of their corresponding HR samples. This is also enforced at the classification stage via the targeted reconstruction loss, which promotes similar features for HR and VLR samples of the same class. Since the base architecture of the proposed model is a capsule network, we first briefly explain its functioning, followed by the in-depth explanation of the proposed model.

\subsection{Preliminaries: Capsule Networks}
\label{sec:preliminaries}
Hinton \textit{et al.} \cite{hinton2011transforming} proposed the concept of \textit{capsules} as an effective method of learning representations. It was further developed by Sabour \textit{et al.} \cite{capsules}, where a capsule network (CapsNet) is presented for classification. A capsule is a ``group of neurons whose activity vector represents the instantiation parameters of a specific type of entity such as an object or an object part''. In other words, instead of a single scalar output, each capsule outputs a vector, the values of which are referred to as the activity vector. The length of each capsule vector ($\|.\|_2$) is bounded in the range of $[0-1]$. Sabour \textit{et al.} \cite{capsules} proposed the concept of dynamic routing between capsules, wherein multiple layers of capsules were stacked for object classification. The final layer contains the classification capsules of dimension $k\times m$, where $k$ is the number of classes and $m$ is the capsule dimension. For a given input, the predicted class is the class corresponding to the capsule with the maximum activity vector (length). In order to learn an effective classification model, margin loss is used to learn the network. Given a $K$ class problem, with $v_k^{x^c}$ as the output of the $k^{th}$ class capsule for an input $x^c$ (belonging to class $c$), and $T_k$ being the label corresponding to the $k^{th}$ class, the margin loss of CapsNet is defined as: 
\begin{equation}
\label{eq:margin}
\begin{split}
    \mathcal{L}_{Margin} = & \sum_{k=1}^{K} \big(T_k\max(0, m^{+} - \|v_k^{x^c}\|)^2 \ \\
    & + \ \lambda(1 \ - \ T_k)\max(0, \|v_k^{x^c}\| - m^{-})^2\big)
\end{split}
\end{equation}
where, $T_k\in\{0,1\}$, that is, whether the input sample belongs to class $k$ ($T_k=1$) or not ($T_k=0$). $m^{+}$ and $m^{-}$ correspond to the positive and negative margin used to increase the intra-class similarity and reduce the inter-class similarity, respectively, and $\lambda$ is a constant for controlling the weight of each term. The above loss (Equation \ref{eq:margin}) promotes a larger length of capsule ($\|v_k\|$) for the correct class, and a smaller length for capsules corresponding to the other classes. Capsule networks are relatively less explored in the literature, with limited or no modification to the architecture or loss function. They have been used for brain tumor detection \cite{brain}, sea grass detection \cite{seagrassCapsule}, generating synthetic data \cite{capsuleGan18}, and image classification \cite{capsnet18}. Capsule networks encode the instantiation parameters for a given input, and thus present the potential of being the appropriate network for VLR image recognition. 

\begin{comment}
\begin{figure}
\centering
\includegraphics[width=3.2in]{sampleSVHN.pdf}
\caption{Sample images from the SVHN dataset \cite{svhn} demonstrating limited information content in VLR images ($8\times8$) as compared to their corresponding HR images ($32\times32$). }
\label{fig:samples}
\end{figure}
\end{comment}

\begin{figure*}
\centering
\includegraphics[width=6.7in]{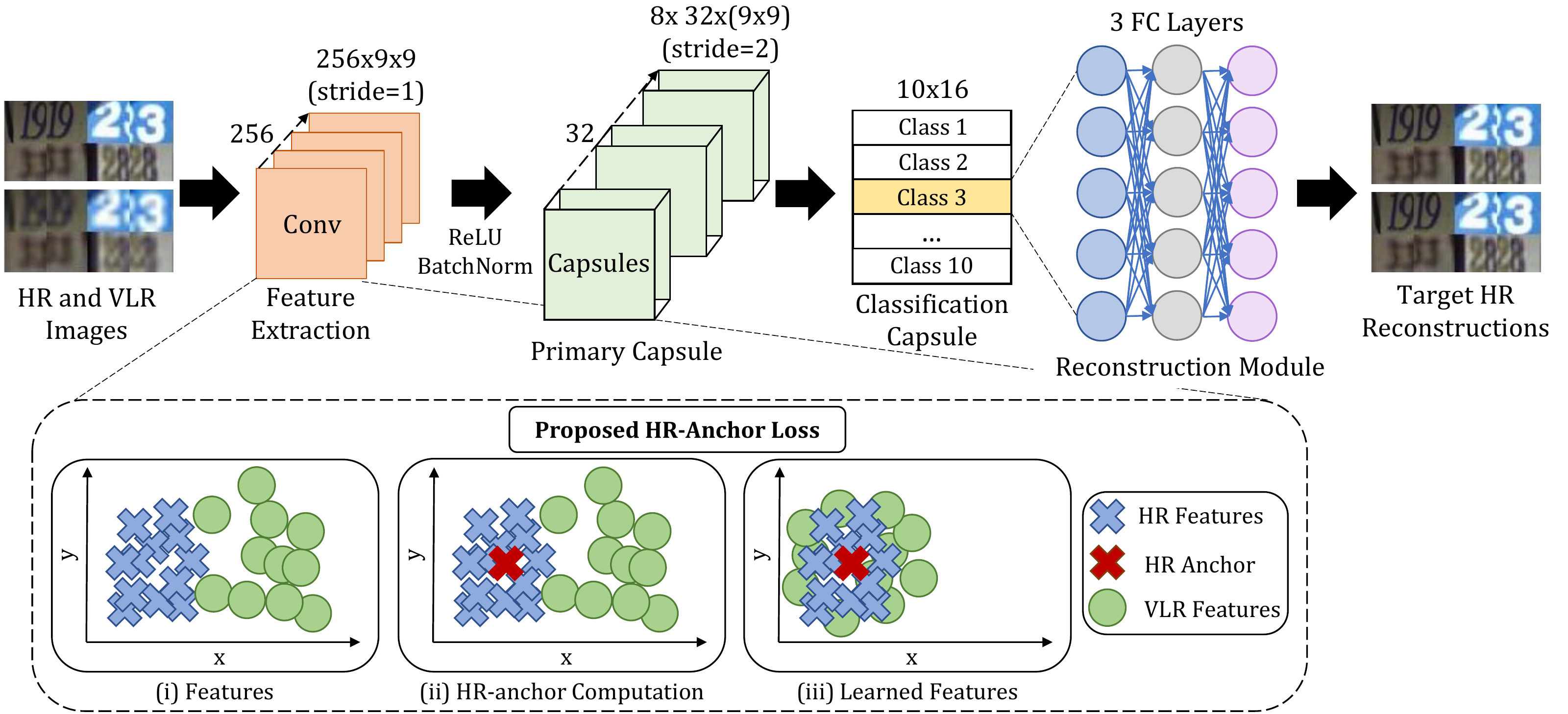}
\caption{Architecture of the proposed Dual Directed Capsule Network (DirectCapsNet) for the SVHN dataset \cite{svhn}. A diagrammatic representation of the HR-Anchor loss is presented for a given class. HR images are used to complement the features learned by the VLR recognition model by directing the model to learn discriminative and information rich features.}
\label{fig:DirectCapsNet}
\vspace{-10pt}
\end{figure*}

\subsection{Proposed DirectCapsNet}

As shown in Figure \ref{fig:DirectCapsNet}, the proposed DirectCapsNet network can be broken down into three components: (i) input, (ii) feature extraction, and (iii) classification. At the time of training, the input consists of both HR and VLR samples. The feature extraction module consists of convolutional layers and the proposed HR-anchor loss, and the classification module consists of a capsule network coupled with the proposed targeted reconstruction loss. By enforcing dual direction via the proposed (i) HR-anchor loss and (ii) targeted reconstruction, the proposed DirectCapsNet focuses on learning meaningful feature-rich representations for VLR inputs, aided by the auxiliary HR samples. The loss function of the proposed DirectCapsNet is formulated as:
\begin{equation}
\begin{gathered}
\label{eq:finalShort}
\mathcal{L}_{DirectCapsNet} = \mathcal{L}_{Margin} \ + \lambda_1\mathcal{L}_{HR-anchor} \\
+ \lambda_2\mathcal{L}_{T-Recon}
\end{gathered}
\end{equation}
where, $\lambda_1$ and $\lambda_2$ are used to balance the weights of the HR-anchor and targeted reconstruction loss with respect to the margin loss. The margin loss introduces discriminability between classes, while the HR-anchor loss and targeted reconstruction loss enforce information-rich representations at the feature and classification level. At the time of testing, for a given VLR input, the class capsule with the highest length is chosen as the class of the given input. It is essential to note that, simulating the real world scenarios, DirectCapsNet utilizes the HR samples only at the time of training, and operates with a given VLR image during testing. As will be demonstrated in the remainder of this section, each component of the proposed model facilitates learning discriminative features for VLR recognition.

\noindent\textbf{Proposed HR-anchor Loss:} Input samples in Figure \ref{fig:DirectCapsNet} are HR ($32\times32\times3$) and VLR ($8\times8\times3$ resolution upscaled to HR resolution) images from the SVHN dataset \cite{svhn}. The limited information content in VLR images makes it difficult to extract discriminative information, often resulting in ineffective recognition, a phenomenon observed in humans as well \cite{sinha2006face}. The proposed HR-anchor loss addresses this challenge by pushing VLR features closer to their HR counter parts. This ensures learning of a discriminative space for VLR recognition, even with limited information. For an input $x^c$ belonging to class $c$, with features $f^{x^c}$ learned from the convolutional layers, the HR-anchor loss is formulated as:
\begin{equation}
\label{eq:HRassisted}
\mathcal{L}_{HR-anchor} = \frac{1}{2}\big((1-r^{x^c})\|f^{x^c} - \mathbb{A}^c \|_2^2 \ + \ r^{x^c}\ \|f^{x^c} - A^{c} \|_2^2\big) \
\end{equation}
where, $r^{x^c}$ is a binary variable to denote the resolution of the sample, i.e., $r^{x^c}=1$ for a HR sample, and $r^{x^c}=0$ for a VLR sample. Since HR samples are only used during training, this information is readily available. $f^{x^c}$ refers to the features extracted from the convolutional layers in the feature module, $A^c$ and $\mathbb{A}^c$ both refer to the HR-anchor of class $c$, which is used to enhance the VLR representations. Specifically, $\mathbb{A}^c$ refers to the HR-anchor in a constant state, whereas ${A}^c$ represents the HR-anchor in a parameter form, which needs to be optimized. The HR-anchor of a particular class corresponds to the average feature vector of all HR samples belonging to that class. Given a VLR sample ($r^{x^c}=0$), the first part of Equation \ref{eq:HRassisted} ($\|f^{x^c} - \mathbb{A}^c \|_2^2$) is active, where the HR-anchor of class $c$ assists the VLR feature $f^{x^c}$ to be closer to the anchor, thereby facilitating learning of discriminative features useful for classification. For a HR sample ($r^{x^c}=1$), the second half of Equation \ref{eq:HRassisted} ($\|f^{x^c} - {A}^c \|_2^2$) becomes active, where both the HR-anchor and features are updated. 

The proposed HR-anchor loss is a combination of learning the HR-anchors and learning the VLR features closer to the HR feature space, in order to learn discriminative VLR features. The first term attempts to direct the VLR features towards the HR anchors, and the second term learns representative HR anchors from the HR features. It is important to note that there is no contribution of the VLR features in the anchor generation, since the HR anchors are constant in the first term. This ensures that the VLR features are directed towards the higher quality HR features, and not the other way round. Therefore, Equation \ref{eq:HRassisted} promotes the learning of informative VLR features with assistance from the HR samples. \\
%In the DirectCapsNet, the input to Equation \ref{eq:HRassisted} are the features learned from the convolutional layers of the feature extraction module (Figure \ref{fig:DirectCapsNet}). \\

\noindent\textbf{Proposed Targeted Reconstruction Loss:} The second form of direction is imposed via the targeted reconstruction loss (Figure \ref{fig:DirectCapsNet}) at the classification module (capsule network). The targeted reconstruction loss promotes learning similar classification capsules for HR and VLR samples.  %The third component of the proposed Dual Directed CapsNet (Figure \ref{fig:DirectCapsNet}) is the classification module, where a capsule network is used as the backbone. 
As explained previously, a capsule is a vector which encodes the instantiation parameters of the input sample \cite{capsules}. For a given input, the activations of a capsule are termed as the activity vector. For reconstruction, only the activity vector of the target class is selected and used to reconstruct the input sample. For an input image $x^c$ belonging to class $c$, the reconstruction loss is mathematically formulated as:
\begin{equation}
\label{eq:recon}
\mathcal{L}_{Recon} = \frac{1}{2} \|x^c - g(v_c^{x^c}) \|_2^2
\end{equation}
where, $v_c^{x^c}$ is the activity vector of the classification capsule of the $c^{th}$ class for the input $x^c$, and $g(.)$ refers to the reconstruction network. The reconstruction loss attempts to encode instantiation parameters that are able to explain the input image, and thus are able to reconstruct the input. Intuitively, we believe that the instantiation parameters of a HR sample and its corresponding VLR sample should be similar. Therefore, in order to incorporate a second level of direction, the targeted reconstruction loss is introduced in the proposed DirectCapsNet.

The targeted reconstruction loss enforces the HR counter-part of a VLR image at the output of the reconstruction network. Regardless of a HR or a VLR input, the reconstructed sample is forced as a HR image. %Therefore, for a HR input, the same is reconstructed at the output of the reconstruction network, whereas the HR counter-part is used for the VLR input. 
For an input $x^c$, the targeted reconstruction loss can be written as:
\begin{equation}
\label{eq:t-recon}
\mathcal{L}_{T-Recon} = \|hr^{x^c} - g(v^{x^c}_c) \|_2^2
\end{equation}
where, $hr^{x^c}$ is the HR image corresponding to the input HR/VLR sample and $v_c^{x^c}$ is the activity vector of the $c^{th}$ class. In case of a HR input image, Equation \ref{eq:t-recon} ensures that the HR input is reconstructed at the output of the reconstruction network. For a VLR image, its HR counter-part is provided as the target of the reconstruction network. Since the reconstruction network operates on the final classification capsule, the targeted reconstruction loss pushes the HR and VLR samples to have a similar capsule activity vector, driven by the HR samples. Therefore, the reconstruction loss promotes learning similar capsule features for HR and VLR samples directly at the classification stage, by directing the model to reconstruct a HR sample from an extracted VLR feature.

Equations \ref{eq:HRassisted} and \ref{eq:t-recon} are combined to update Equation \ref{eq:margin} and the loss function of the proposed DirectCapsNet for an input $x^c$ (belonging to class $c$) is written as:
\begin{equation}
\begin{gathered}
\label{eq:final}
\mathcal{L}_{DirectCapsNet} = \sum_{k=1}^{K} \Big(T_k\max(0, m^{+} - \|v_k^{x^c}\|)^2 + \ \\
\lambda\ (1-T_k)\ \max(0, \ \|v_k^{x^c}\| - m^{-})^2\Big) \ + \ \frac{1}{2}\Big(\lambda_1 (1-r^{x^c}) \\
\ \|f^{x^c} - \mathbb{A}^c \|_2^2 \ +\lambda_1r^{x^c}\ \|f^{x^c} - A^{c} \|_2^2 \ + \lambda_2\ \|hr^{x^c} - g(v^{x^c}_c) \|_2^2\Big)
\end{gathered}
\end{equation}
%The above model \textit{directs} the VLR features towards their feature-rich HR counter-parts, thus aiding VLR recognition. 

\subsection{Implementation Details}
DirectCapsNet has been implemented in Python, using the PyTorch framework on the NVIDIA Tesla P-100 GPU. Adam optimizer \cite{adam} has been used for learning the model. The weight of the HR-anchor loss ($\lambda_1$ of Equation \ref{eq:final}) is set to $1e-3$, and the weight of the targeted reconstruction loss ($\lambda_2$ of Equation \ref{eq:final}) is set to $1e-5$. The positive and negative margins for the margin loss ($m^{+}$ and $m^{-}$ of Equation \ref{eq:margin}) are set to 0.9 and 0.1, respectively. As shown in Figure \ref{fig:DirectCapsNet}, for all the experiments, the DirectCapsNet model contains $n$ convolution layers, followed by two capsule layers. The HR-anchor loss is applied on the final convolution layer of the DirectCapsNet. The final capsule layer is connected to a reconstruction network of three fully connected layers. For cases where the HR samples are larger than $96\times96$, three convolutional layers with $[16, 32, 128]$ filters are used with a batch size of 32 samples. In cases where the HR samples are smaller, a convolutional layer with 128 filters is used with a batch size of 100 samples. $ReLU$ activation function is used between the convolutional layers along with batch normalization \cite{batch}. All models have been trained from scratch and no pre-trained networks have been used. % Model files of the proposed DirectCapsNet will be released for reproducibility of results.

\section{Experiments and Protocols}
The proposed DirectCapsNet has been evaluated for three very low resolution (VLR) recognition problems: (i) VLR digit recognition, (ii) VLR face recognition, and (iii) unconstrained VLR face recognition. Details regarding the dataset and protocols for each case study are as follows:

\noindent \textbf{Case study 1 - VLR Digit Recognition:} The Street View House Numbers (SVHN) dataset \cite{svhn} has been used for VLR digit recognition. The dataset contains real-world images of digits in the range $[0-9]$. Pre-defined benchmark protocol has been used for the given 10-class problem, wherein 73,257 digits are used for training and 26,032 digits are used for testing. For VLR recognition, consistent with the existing protocol \cite{wang2016studying}, $32\times32$ HR images are used, and $8\times8$ VLR images are used. Results are reported in terms of the top-1 and top-5 accuracies.

\noindent \textbf{Case study 2 - VLR Face Recognition:} VLR face recognition has direct applicability in scenarios of image tagging or situations where multiple people are captured in a single image. %VLR face recognition has wide applicability in surveillance scenarios, where an image to be recognized is of low-resolution. 
For this particular case-study, experiments have been performed on the CMU Multi-PIE dataset \cite{mpie} which simulates a constrained setting. Consistent with the existing protocol \cite{singh18CvprW}, 237 subjects are used. One image per subject is added to the training set/gallery which consists of the HR images, and one image per subject is added to the testing set/probe (VLR). The HR images are of $96\times 96$ resolution and the VLR images are of $8\times8$ and $16\times16$, respectively. Results are reported using the rank-1 identification accuracy. 

\noindent \textbf{Case study 3 - Unconstrained VLR Face Recognition:} Unconstrained VLR face recognition has wide applicability in surveillance scenarios, where the VLR face image often contains other variations such as pose, illumination, and occlusion. Experiments have been performed on two datasets: (a) UnConstrained College Students (UCCS) dataset \cite{uccs} for an unconstrained surveillance setting and (b) CMU Multi-PIE dataset \cite{mpie} with pose and illumination variations for a semi-constrained setting.  

\noindent The UCCS dataset contains images of college students, captured using a long-range high-resolution surveillance camera kept at a standoff distance of 100 to 150 meters. The images show students walking around the campus, between classes. The large standoff distance and unconstrained nature of the data simulates real world surveillance settings. The dataset contains a labeled subset of 1732 identities. Consistent with the existing protocol \cite{ge19Tip,wang2016studying}, a subset containing the top 180 identities (in terms of the number of images) is used for evaluation. As per the protocol, each subject's images are divided into a $4:1$ ratio corresponding to training:testing. The VLR images are of $16\times16$ resolution, whereas the HR images are of $80\times80$ pixels.

\noindent As described above, CMU Multi-PIE dataset \cite{mpie} contains images with pose, expression, and illumination variations. As per the existing protocol \cite{mudunuriPami16}, in this case-study face recognition is performed across pose and illumination variations for VLR images. Images pertaining to 50 subjects are used for training and images of the remaining subjects form the test set. In our experiments, we do not utilize the training set and only use the gallery images of the test set in order to train the proposed DirectCapsNet model. The gallery comprises of the frontal images (used for training the proposed model), and the probe (test set) are images having a different pose (`05\_0' of the dataset) and illumination. Experiments are performed across five different pairs of illumination conditions and average rank-1 identification accuracy has been reported. Consistent with \cite{mudunuriPami16}, the HR images are of $36\times30$ resolution, while VLR images have resolution of $18\times15$, $15\times12$, $12\times10$, and $10\times9$.

%The HR images are down-sampled to generate the VLR images, which are then bicubicly interpolated to the HR image's dimension.
Figure \ref{fig:vlr_hr} presents some HR and VLR images from the datasets used in the three case-studies. Bicubic interpolation is used for conversion from HR to VLR and vice-versa. At the time of training, the HR and VLR pairs are used for the targeted reconstruction loss. Data augmentation is applied by introducing brightness variations, flipping along the y-axis, and random crops. At the time of testing, only the VLR image is provided for classification. 

\section{Results and Analysis}
\begin{figure}
\centering
\includegraphics[width=3.2in]{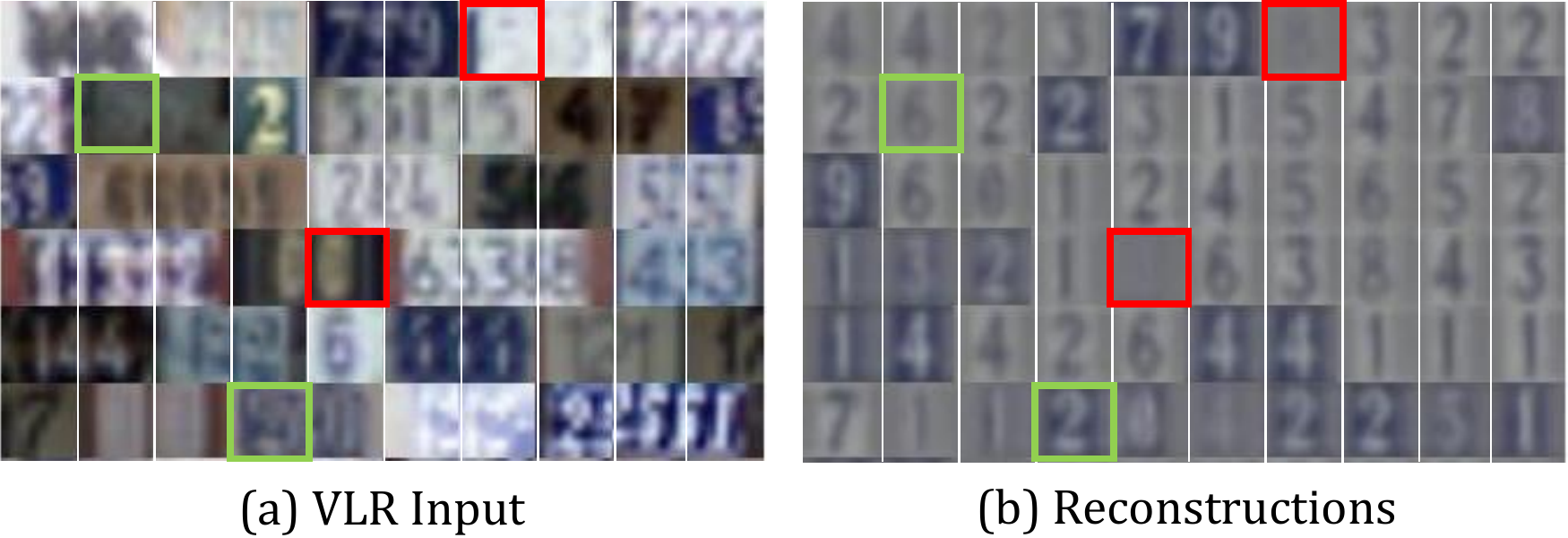}
\caption{Sample reconstructions obtained on the SVHN dataset from VLR input. DirectCapsNet is able to reconstruct digits where limited information content is available (e.g. green boxes), however it also fails to correctly reconstruct some challenging cases (e.g. red boxes).}
\vspace{-10pt}
\label{fig:recon}
\end{figure}

\begin{figure}
\centering
\includegraphics[width=3.2in]{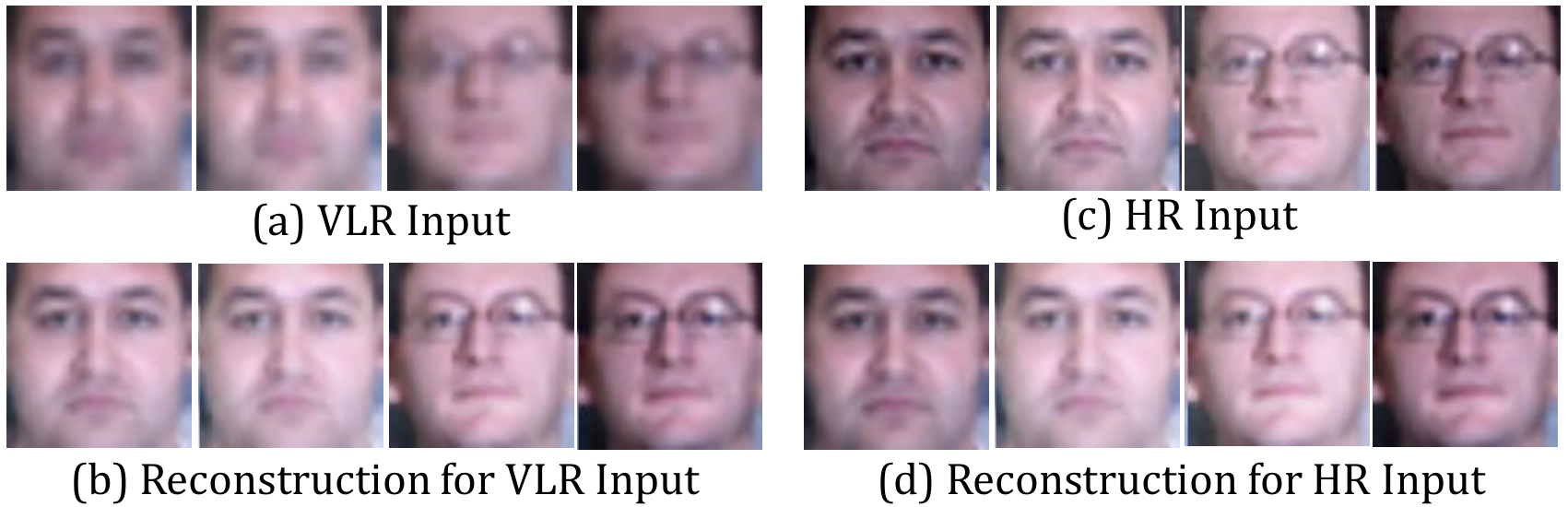}
\caption{Sample reconstructions obtained from the proposed DirectCapsNet model on the CMU Multi-PIE dataset. For the same class, DirectCapsNet is able to project VLR and HR samples onto a similar target, suggesting robust resolution-invariant feature representations.}
\label{fig:reconFace}
\end{figure}

\begin{figure}
\centering
\includegraphics[width=3in]{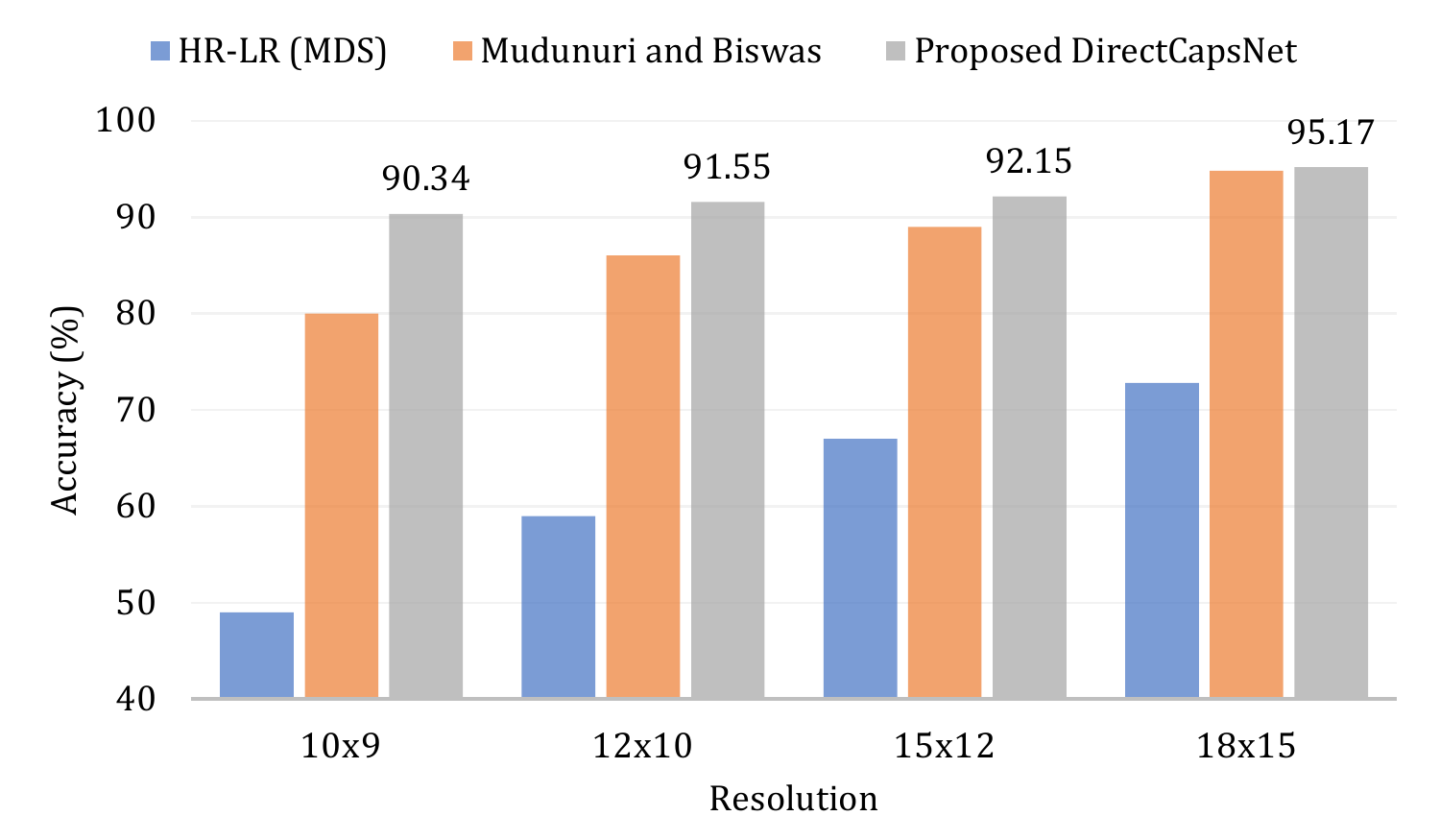}
\caption{Performance of the proposed DirectCapsNet for varying resolutions of VLR face recognition with pose and illumination variations. The HR resolution was fixed to $36\times30$ pixels. Comparison has been shown with HR-LR (MDS) \cite{mds} and Mudunuri and Biswas \cite{mudunuriPami16}.}
\label{fig:graph}
\end{figure}

Tables \ref{tab:svhn} - \ref{tab:uccs} and Figures \ref{fig:recon} - \ref{fig:graph} present the results for the three case-studies: (i) VLR digit recognition, (ii) VLR face recognition, and (iii) unconstrained VLR face recognition. Analysis of the proposed DirectCapsNet has also been performed in order to demonstrate the effectiveness of each component. Since existing protocols have been used for analysis, results have directly been reported from the respective publications. 

\begin{table}
\centering
\caption{Top-1 and top-5 accuracy (\%) on the SVHN dataset \cite{svhn} for VLR digit recognition ($8\times8$). }
\label{tab:svhn}
\begin{tabular}{|c|l|c|c|}
\hline
& \multirow{2}{*}{\textbf{Algorithm}} & \multicolumn{2}{c|}{\textbf{Accuracy (\%)}}\\
\cline{3-4}
& & Top-1 & Top-5 \\
\hline
\hline
& CNN (VLR) (2016) \cite{wang2016studying} & 45.29 & 66.78 \\
\cline{2-4}
& RPC Nets (2016) \cite{wang2016studying} & 56.98 & 70.82 \\
\hline
\hline
 \parbox[t]{2mm}{\multirow{5}{*}{\rotatebox[origin=c]{90}{Proposed}}} &
CapsNet (HR) & 77.82 & 87.86\\
\cline{2-4}
& CapsNet (VLR) & 79.19 & 88.89\\
\cline{2-4}
& DirectCapsNet - (HR-anchor Loss) & 82.42 & 90.15\\
\cline{2-4}
& DirectCapsNet - (Targeted Recon.) & 81.95 & 90.35\\
\cline{2-4}
\cline{2-4}
& \textbf{Proposed DirectCapsNet} & \textbf{84.51} & \textbf{91.20} \\
\hline
\end{tabular}
\end{table}

\noindent \textbf{Ablation Study and Analysis of DirectCapsNet:} Experiments have been performed on the SVHN dataset to analyze each component of the proposed DirectCapsNet, and motivate their inclusion in the final model. As observed from Table \ref{tab:svhn}, the native CapsNet model (having the margin loss) when trained on VLR images (CapsNet (VLR)) attains the top-1 classification accuracy of 79.19\%, which demonstrates large improvement over the native CNN architecture (45.29\%) \cite{wang2016studying}. The improved performance promotes the usage of capsule networks for the task of VLR recognition. Consistent with literature \cite{capsules}, it is our belief that since CapsNet attempts to encode the instantiation parameters of the data, it results in learning features invariant to minor variations, a desirable property of a robust VLR recognition module. 

Further, in order to reaffirm the necessity of a VLR recognition model, a CapsNet with the same architecture is trained on HR images only. In this case, the model does not see any VLR images at the time of training and is evaluated on VLR test images. As can be observed, the CapsNet (HR) achieves a classification accuracy of 77.82\%, thus reaffirming the need to develop dedicated VLR recognition networks or utilize task-specific information while training. We also performed the McNemar test \cite{mcnemar} and achieved statistical difference at a confidence interval (C.I.) of 99\% ($p$-value$<$0.01) between the proposed DirectCapsNet and CapsNet. Table \ref{tab:svhn} can also be analyzed to understand the effect of each component of the proposed DirectCapsNet model. Upon removing the HR-anchor loss from the DirectCapsNet model, top-1 accuracy of 82.42\% is achieved, whereas, removal of the targeted reconstruction loss results in a top-1 accuracy of 81.95\%. Both these models demonstrate poor performance as compared to the proposed DirectCapsNet model, thus supporting the inclusion of the HR-anchor loss, targeted reconstruction loss, and capsules in the DirectCapsNet model. \\

\noindent \textbf{Case study 1 - VLR Digit Classification:} Table \ref{tab:svhn} presents the top-1 and top-5 classification accuracy for the SVHN dataset of the proposed DirectCapsNet and comparison with other techniques. The proposed DirectCapsNet model achieves top-1 accuracy of 84.51\% and top-5 accuracy of 91.20\%. DirectCapsNet demonstrates an improvement of over 27\% at top-1 with respect to the state-of-the-art results of Robust Partially Coupled Networks (RPC Nets) \cite{wang2016studying}, which is a CNN based framework to learn partial shared weights for VLR and HR samples, and partial independent weights for the two. The superior performance of the proposed DirectCapsNet model motivates its usage for VLR recognition. Figure \ref{fig:recon} presents sample reconstructions obtained from the DirectCapsNet for $8\times8$ VLR samples. It is motivating to note that the DirectCapsNet model is able to reconstruct the digits for the input samples, which motivates the inclusion of the targeted reconstruction loss. Similar reconstructions are obtained for samples of the same class, which demonstrate the effectiveness of the HR-anchor loss for increasing the intra-class similarity between features. 

\begin{table}
\centering
\caption{Rank-1 accuracy (\%) for VLR recognition on the CMU Multi-PIE dataset \cite{mpie}. The HR images are of $96\times96$ resolution.}
\label{tab:mpieLR}
\begin{tabular}{|l|c|c|}
\hline
\multirow{2}{*}{\textbf{Algorithm}} & \multicolumn{2}{c|}{\textbf{Accuracy (\%)}}\\
\cline{2-3}
& $8\times8$ & $16\times16$ \\
\hline
\hline
Original + COTS (2018) \cite{singh18CvprW} & 0.0 & 0.0 \\
\hline
Bicubic Interp. + COTS (2018) \cite{singh18CvprW} & 0.1 & 1.1 \\
\hline
SHSR (Synthesis + COTS) (2018) \cite{singh18CvprW} & 82.6 & 91.1 \\
\hline
\hline
\textbf{Proposed DirectCapsNet} & \textbf{94.5} & \textbf{97.4} \\
\hline
\end{tabular}
\end{table}

\begin{table}
\centering
\caption{Rank-1 accuracy (\%) on the UCCS dataset \cite{uccs} for VLR face recognition ($16\times16$). The HR images are of $80\times80$ resolution.}
\label{tab:uccs}
\begin{tabular}{|l|c|}
\hline
\textbf{Algorithm} & \textbf{Acc. (\%)}\\
\hline
\hline
Robust Partially Coupled Nets (2016) \cite{wang2016studying} & 59.03 \\
\hline
Selective Knowledge Distillation (2019) \cite{ge19Tip} & 67.25 \\
\hline
LMSoftmax for VLR (2019) \cite{li19Tifs} & 64.90 \\
\hline
L2Softmax for VLR (2019) \cite{li19Tifs} & 85.00  \\
\hline
Centerloss for VLR (2019) \cite{li19Tifs} &  93.40 \\
\hline
\hline
\textbf{Proposed DirectCapsNet} & \textbf{95.81} \\
\hline
\end{tabular}
\end{table}

\begin{figure}
\centering
\includegraphics[trim={0.5in, 0in, 0.5in, 0in}, clip, width=3.2in]{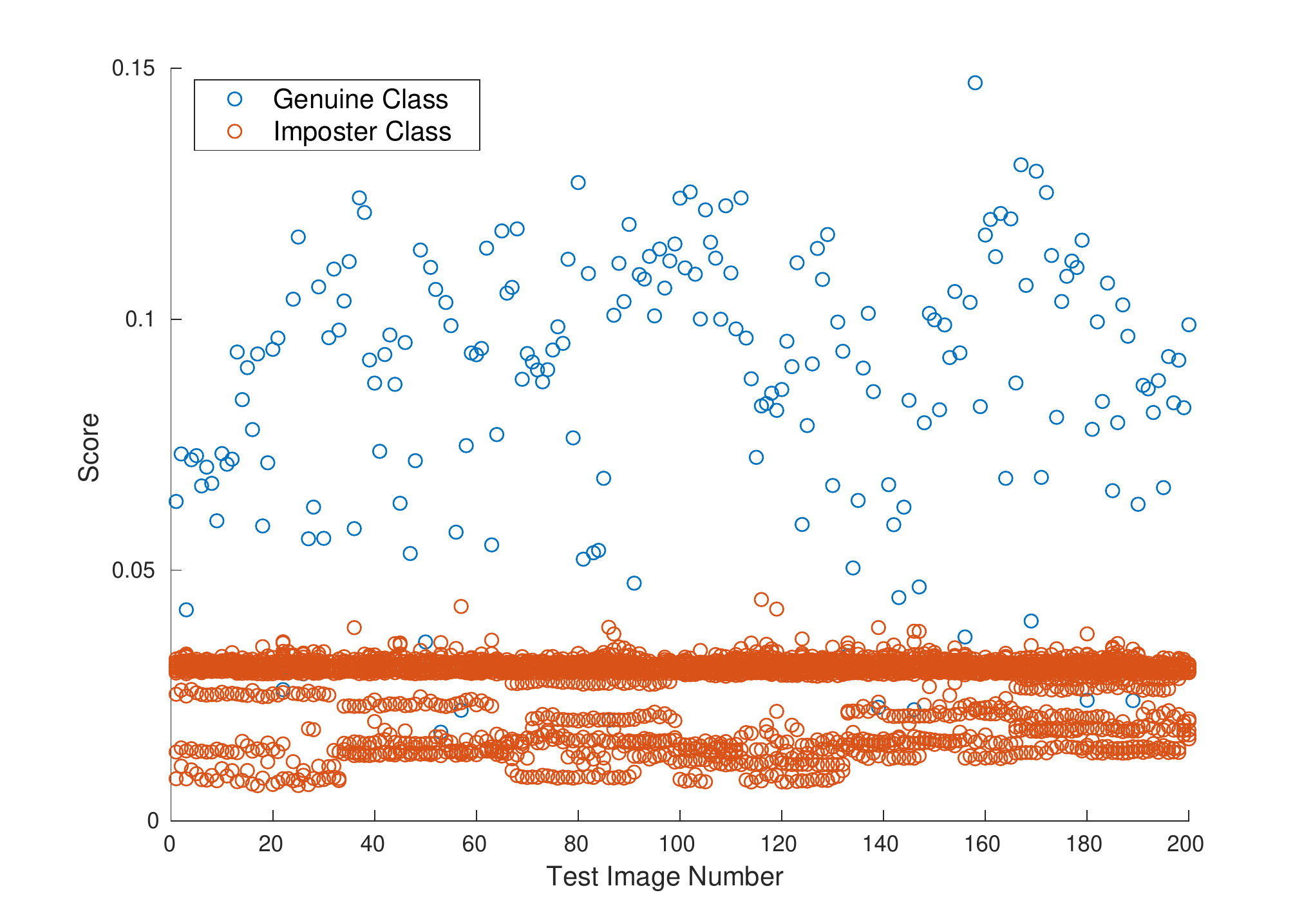}
\caption{Scores obtained by the proposed DirectCapsNet for VLR recognition on some samples of the UCCS dataset. Each test image has one genuine score (correct class) and 179 imposter scores (incorrect class).}
\label{fig:score}
\vspace{-10pt}
\end{figure}

\noindent \textbf{Case study 2 - VLR Face Recognition:} Table \ref{tab:mpieLR} presents the rank-1 identification (or top-1 recognition) accuracy for two protocols of VLR face recognition. The proposed DirectCapsNet model achieves an accuracy of 94.5\% and 97.4\% for $8\times8$ and $16\times16$ VLR images, while having the HR auxiliary images as $96\times96$ (Table \ref{tab:mpieLR}) on the constrained CMU Multi-PIE dataset. DirectCapsNet demonstrates an improvement of almost 12\% as compared to the state-of-the-art (Synthesis via Hierarchical Sparse Representations (SHSR)) \cite{singh18CvprW} for $8\times8$ resolution images. Figure \ref{fig:reconFace} presents sample VLR and HR face images, along with the reconstructions obtained from the DirectCapsNet. The proposed model is able to reconstruct faces belonging to the same subject onto a similar target, suggesting high within-class similarity. Both VLR and HR samples are reconstructed as similar images, which reinstates the benefit of the targeted reconstruction and HR-anchor loss.

\begin{comment}
\begin{table}
\centering
\caption{Rank-1 accuracy (\%) for VLR face recognition ($18\times15$) with pose and illumination variations. The HR images were of $36\times30$ resolution. Comparative accuracies have been taken from Mudunuri and Biswas \protect\cite{mudunuriPami16}.}
\label{tab:pose}
\begin{tabular}{|l|c|}
\hline
\textbf{Algorithm} & \textbf{Accuracy (\%)}\\
\hline
\hline
HR-LR (baseline) \cite{mds} & 59.04 \\
\hline
LSML \cite{lsml} & 89.17 \\
\hline
SFRD + PMML \cite{sfrd} & 92.72 \\
\hline
Mudunuri and Biswas \cite{mudunuriPami16} & 94.81 \\
\hline
\hline
\textbf{Proposed DirectCapsNet} & \textbf{95.17} \\
\hline
\end{tabular}
\end{table}
\end{comment}

\noindent \textbf{Case study 3 - Unconstrained VLR Face Recognition:} Table \ref{tab:uccs} and Figure \ref{fig:graph} present the rank-1 identification (or top-1 recognition) accuracy for unconstrained VLR face recognition. As shown in Table \ref{tab:uccs}, on the UCCS dataset, DirectCapsNet model achieves a rank-1 accuracy of 95.81\% demonstrating an improvement of almost 2.5\% over the state-of-the-art network and almost 10\% from the current second best \cite{li19Tifs}. Comparison has also been performed with the recently proposed large-margin softmax (LMSoftmax), $l_2$-constrained softmax (L2Softmax), and center-loss based VLR recognition systems \cite{li19Tifs}. The improved performance of the proposed DirectCapsNet over metric learning techniques demonstrates the benefit of incorporating auxiliary HR information to provide direction while training with the proposed dual directed loss functions. Figure \ref{fig:score} presents the scores obtained on samples of the UCCS dataset by the DirectCaspNet model. The scores correspond to the length of the activity vectors of the capsules used for classification. Figure \ref{fig:score} suggests that the model is able to generate a high score for the correct class and a small score for the other classes, which promotes separability, resulting in high recognition performance.

Similar performance is obtained on the CMU Multi-PIE dataset (Figure \ref{fig:graph}) with pose and illumination variations, where the proposed DirectCapsNet achieves an average recognition performance of 95.17\%, demonstrating an improvement of around 1.64\% from the current state-of-the-art algorithm \cite{mudunuriPami16}. Figure \ref{fig:graph} demonstrates that the proposed DirectCapsNet does not suffer a major decrease in accuracy as other techniques with reducing the resolution. The model achieves the recognition accuracy of 92.15\% and 90.34\% for $15\times12$ and $10\times9$, respectively, whereas, the second best performing model \cite{mudunuriPami16} shows a drop of almost 9\% between the two resolutions. Improved recognition performance across multiple very low resolutions motivates the applicability of the proposed DirectCapsNet model for real world scenarios.

\section{Conclusion}
Existing research has primarily focused on high resolution and low resolution image recognition; however, the problem of VLR recognition has received limited attention. VLR recognition, an arduous problem with wide applicability in real world scenarios, suffers from the primary challenge of low information content. This research presents a novel Dual Directed Capsule Network (DirectCapsNet) for VLR recognition. The DirectCapsNet combines the margin loss for classification with the proposed HR-anchor loss and the targeted reconstruction loss for enhancing the VLR features. HR images are used during training as `auxiliary' data to complement the VLR feature learning. Experimental results on VLR digit recognition (SVHN database) and constrained/unconstrained VLR face recognition (CMU Multi-PIE and UCCS databases) demonstrate the efficacy of the proposed model, and promote its usability for different VLR tasks. In future, we plan to extend the proposed algorithm to address multiple covariates; for example, in face recognition applications, VLR recognition in the presence of disguise \cite{dfw}, aging \cite{aging}, spectral variations \cite{spectral}, and adversarial attacks \cite{adversarial}.

\section{Acknowledgement}
This research is partially supported through the Infosys Center for Artificial Intelligence, IIIT-Delhi, India.  M. Vatsa is also supported through the Swarnajayanti Fellowship by Government of India. S. Nagpal is supported via the TCS PhD fellowship.

{\small
\bibliographystyle{ieee_fullname}
\bibliography{egbib}
}

\end{document}